\documentclass{article}
\usepackage{ijcai20}

\usepackage{times}

\usepackage{soul}
\usepackage{url}
\usepackage[hidelinks]{hyperref}
\usepackage[utf8]{inputenc}
\usepackage[small]{caption}
\usepackage{graphicx}
\usepackage{amsmath}
\usepackage{amsfonts}
\usepackage{booktabs}
\usepackage{subcaption}
\title{Reference-guided Face Component Editing}

\author{
Qiyao Deng$^{1,4}$\and
Jie Cao$^{1,4}$\and
Yunfan Liu$^{1,4}$\and
Zhenhua Chai$^5$\and
Qi Li$^{1,2,4}$\footnote{Contact Author}\and
Zhenan Sun$^{1,3,4}$\\
\affiliations
$^1$Center for Research on Intelligent Perception and Computing,
      NLPR, CASIA, Beijing, China\\
$^2$Artificial Intelligence Research, CAS, Qingdao, China\\
$^3$Center for Excellence in Brain Science and Intelligence Technology,
        CAS, Beijing, China\\
$^4$School of Artificial Intelligence, University of Chinese Academy of
    Sciences, Beijing, China\\
$^5$Vision Intelligence Center, AI Platform, Meituandianping Group\\
\emails
\{qiyao.deng, jie.cao, yunfan.liu\}@cripac.ia.ac.cn,
\{qli, znsun\}@nlpr.ia.ac.cn,
chaizhenhua@meituan.com
}

\begin{document}

\maketitle

\begin{abstract}
Face portrait editing has achieved great progress in recent years. However, previous methods either 1) operate on pre-defined face attributes, lacking the flexibility of controlling shapes of high-level semantic facial components (e.g., eyes, nose, mouth), or 2) take manually edited mask or sketch as an intermediate representation for observable changes, but such additional input usually requires extra efforts to obtain. To break the limitations (e.g. shape, mask or sketch) of the existing methods, we propose a novel framework termed r-FACE (Reference-guided FAce Component Editing) for diverse and controllable face component editing with geometric changes. Specifically, r-FACE takes an image inpainting model as the backbone, utilizing reference images as conditions for controlling the shape of face components. In order to encourage the framework to concentrate on the target face components, an example-guided attention module is designed to fuse attention features and the target face component features extracted from the reference image. Through extensive experimental validation and comparisons, we verify the effectiveness of the proposed framework.
\end{abstract}

\begin{figure}[!tbp]
\setlength{\belowcaptionskip}{-0.4cm}
\setlength{\abovecaptionskip}{0.15cm}
\centering
\includegraphics[width=0.43\textwidth]
{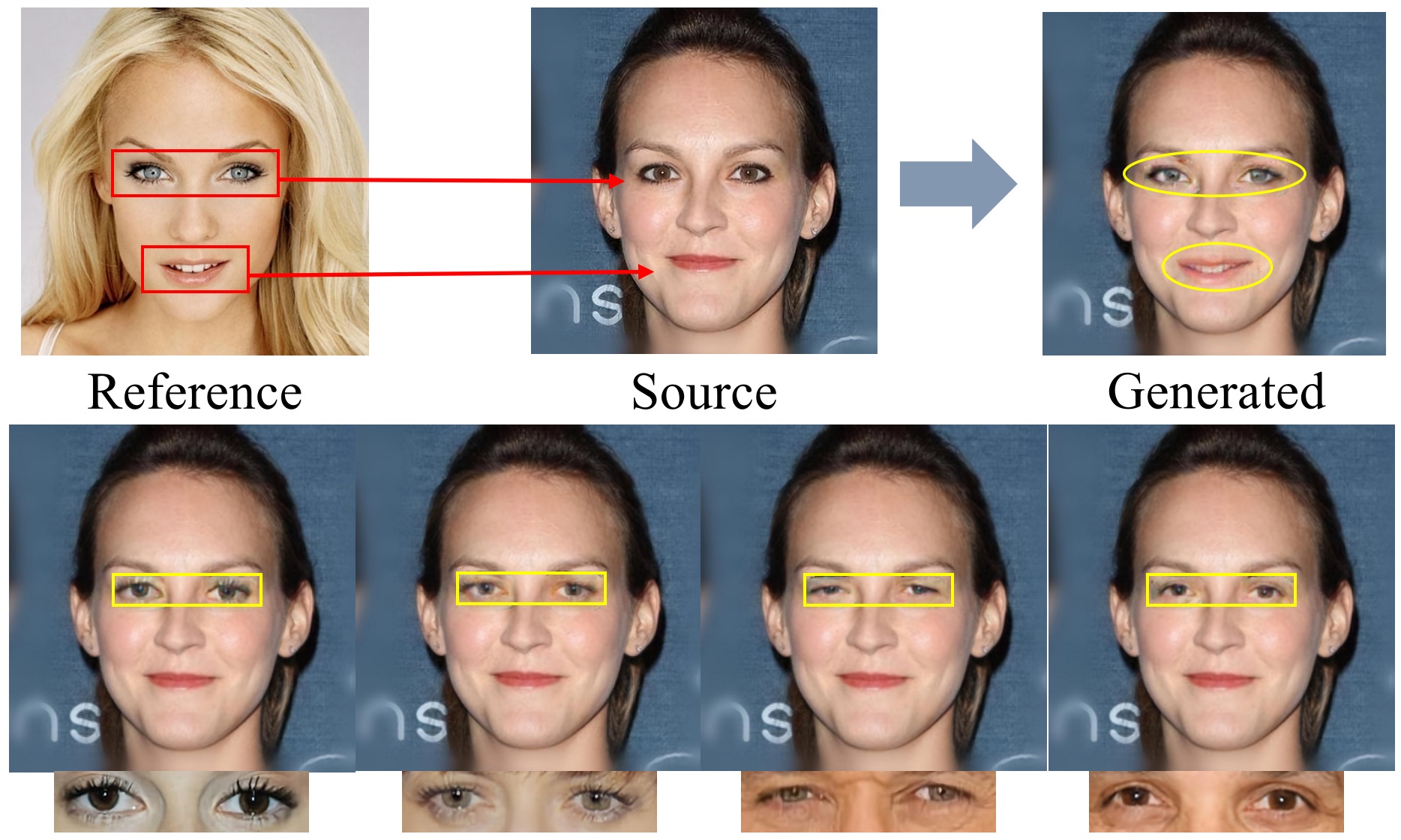}
\caption{The illustration of reference guided face component editing. The first row is the definition diagram of the task, and the second row is the synthesized result based on the given different reference images.}
\label{definition}
\end{figure}
\section{Introduction}
\begin{figure*}[!tbp]
\setlength{\belowcaptionskip}{-0.3cm}
\setlength{\abovecaptionskip}{0.15cm}
\centering
\includegraphics[width=0.88\textwidth]
{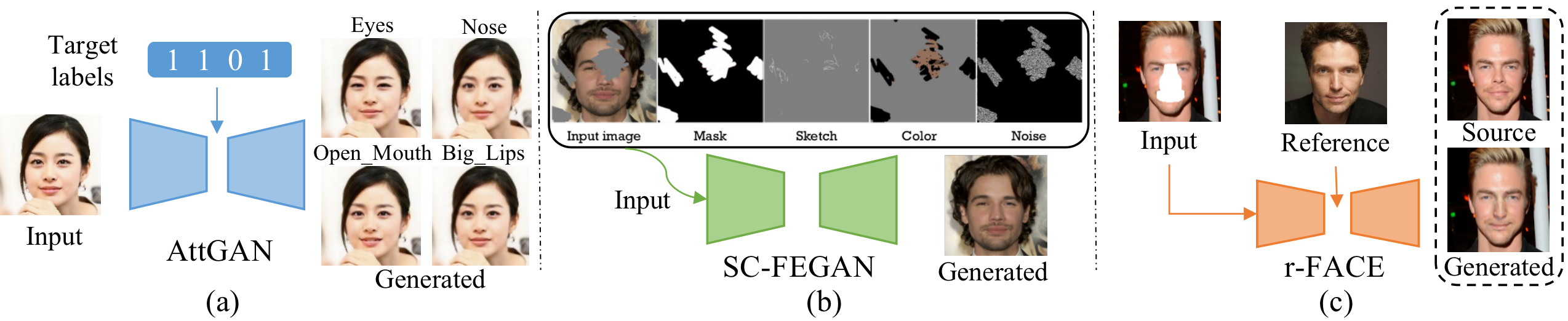}
\caption{Different methods for face portrait editing. (a) AttGAN, (b) SC-FEGAN and (c) Our network.}
\label{ASI}
\end{figure*}
Face portrait editing is of great interest in the computer vision community due to its potential applications in movie industry, photo manipulation, and interactive entertainment, etc.
With advances in Generative Adversarial Networks \cite{goodfellow2014generative} in recent years, tremendous progress has been made in face portrait editing~\cite{yang2018learning,choi2018stargan,liu2019attribute}.
These approaches generally fall into three main categories: label-conditioned methods, geometry-guided methods and reference-guided methods. 
Specifically, label-conditioned methods~\cite{he2019attgan,choi2018stargan} only focus on several pre-defined conspicuous attributes thus lacking the flexibility of controlling shapes of high-level semantic facial components (e.g., eyes, nose, mouth). This is because it is hard to produce results with observable geometric changes merely based on attribute labels. 
In order to tackle this, geometry-guided methods~\cite{Jo_2019_ICCV,gu2019mask} propose to  take manually edited mask or sketch as an intermediate representation for obvious face component editing with large geometric changes. However, directly taking such precise representations as a shape guide is inconvenient for users, which is laborious and requires painting skills. 
To solve this problem, reference-guided methods directly learns shape information from reference images without requiring precise auxiliary representation, relieving the dependence on face attribute annotation or precise sketch/color/mask information. As far as we know, reference-guided methods are less studied than the first two methods. 
ExGANs~\cite{dolhansky2018eye} utilizes exemplar information in the form of a reference image of the region for eye editing (in-painting). However, ExGANs can only edit eyes and requires reference images with the same identity, which is inconvenient to collect in practice. 
ELEGANT~\cite{xiao2018elegant} transfers exactly the same type of attributes from a reference image to the source image by exchanging certain part of their encodings. However, ELEGANT is only used for editing obvious semantic attributes, and could not change abstract shapes.

To overcome the aforementioned problems, we propose a new framework: \textbf{Reference guided FAce Component Editing} (r-FACE for short), which can achieve diverse and controllable face semantic components editing (e.g., eyes, nose, mouth) without requiring paired images. 
The ideal editing is to transfer single or multiple face components from a reference image to the source image, while still preserving the consistency of pose, skin color and topological structure(see Figure~\ref{definition}). Our framework breaks the limitations of existing methods: 1) shape limitation. r-FACE can flexibly control diverse shapes of high-level semantic facial components by different reference images; 2) intermediate presentation limitation. There is no need to manually edit precise masks or sketches for observable geometric changes.

\begin{figure*}[!tbp]
\setlength{\belowcaptionskip}{-0.3cm}
\setlength{\abovecaptionskip}{0.15cm}
\centering
\includegraphics[width=0.85\textwidth]
{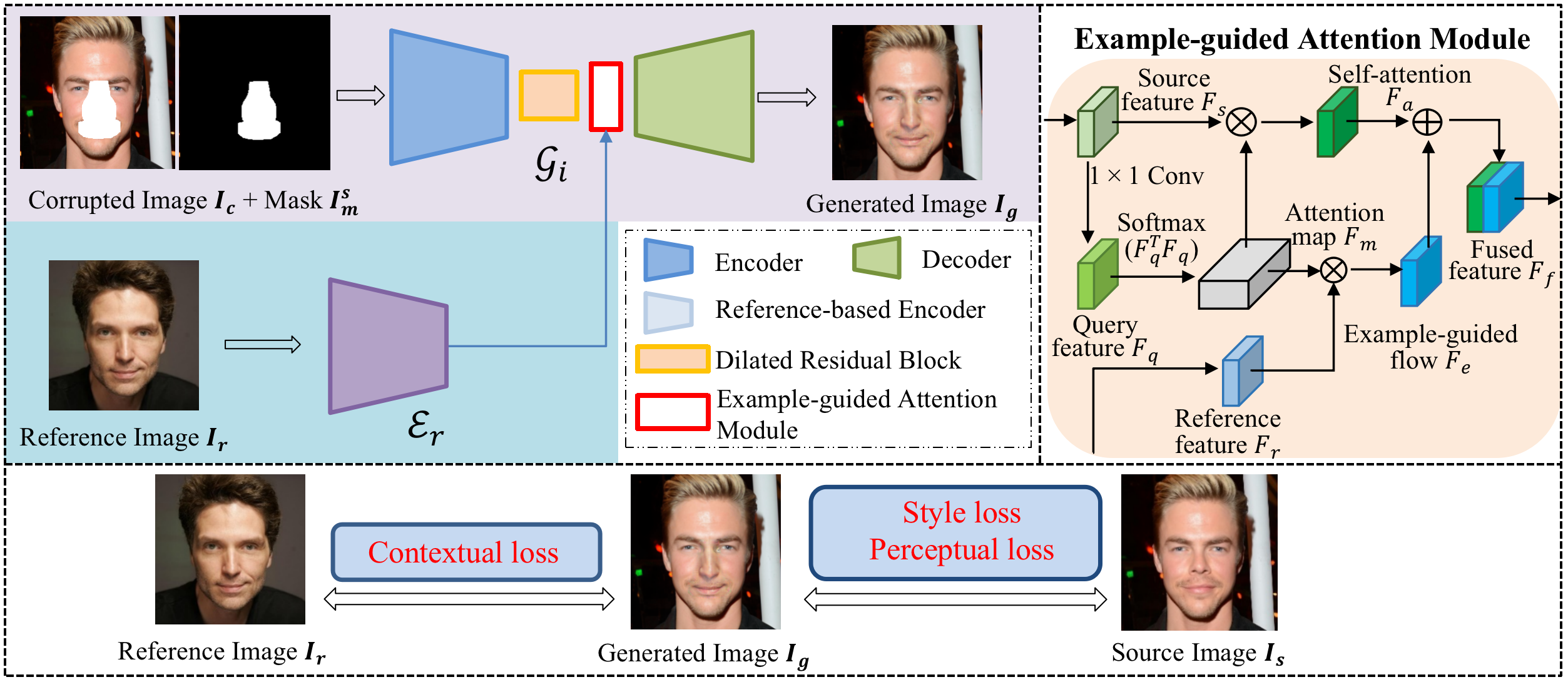}
\caption{The overall structure of proposed framework. On the top left corner is the generator. The top-right figure shows detailed attention module. The constrains among the source image, the reference image and the generated image is shown on the bottom.}
\label{framework}
\end{figure*}

Our framework is based on an image inpainting model for editing face components by reference images even without paired images. r-FACE has two main streams including 1) an inpainting network $\mathcal{G}_{i}$ and 2) an embedding network $\mathcal{E}_{r}$.
As shown in Figure ~\ref{framework}, $\mathcal{G}_{i}$ takes the source image with target face components corrupted and the corresponding mask image as input, and outputs the generated image with semantic features extracted by $\mathcal{E}_{r}$. To encourage the framework to concentrate on the target face components, an example-guided attention module is introduced to combine features extracted by $\mathcal{G}_{i}$ and $\mathcal{E}_{r}$. To supervise the proposed model, a contextual loss is adopted to constrain the similarity of shape between generated images and reference images, while a style loss and a perceptual loss are adopted to preserve the consistency of skin color and topological structure between generated images and source images. Both qualitative and quantitative results demonstrate that our model is superior to existing literature by generating high-quality and diverse faces with observable changes for face component editing.

In summary, the contributions of this paper are as follows:
\begin{itemize}
    \item We propose a novel framework named reference guided face component editing for diverse and controllable face component editing with geometric changes, which breaks the shape and intermediate presentation (e.g., precise masks or sketches) limitation of existing methods.
    \item An example-guided attention module is designed to encourage the framework to concentrate on the target face components by combining attention features and the target face component features of the reference image, further boosting the performance of face portrait editing.
    \item Both qualitative and quantitative results demonstrate the superiority of our method compared with other benchmark methods.
\end{itemize}

\section{Related Work}
\paragraph{Face Portrait Editing. }
Face portrait editing aims at manipulating single or multiple attributes or components of a face image towards given conditions. Depending on different conditions, face portrait editing methods can be classified into three categories: label-conditioned methods, geometry-guided methods and reference-guided methods. Label-conditioned methods change predefined attributes, such as hair color~\cite{choi2018stargan}, age~\cite{liu2019attribute} and pose~\cite{cao20193d}.
However, these methods focus on several conspicuous attributes~\cite{liu2015deep,langner2010presentation}, lacking the flexibility of controlling the shapes of different semantic facial parts. As shown in Figure ~\ref{ASI}(a), AttGAN~\cite{he2019attgan} attempts to edit attributes with shape changes, such as $'Narrow\_Eyes'$, $'Pointy\_Nose'$ and $'Mouth\_Slightly\_Open'$, but it can only achieve subtle changes hard to be observed. Moreover, lacking of labeled data will extremely limit the performance of these methods.
To tackle above problems, geometry-guided methods use an intermediate representation to guide observable shape changes.
~\cite{gu2019mask} proposes a framework based on mask-guided conditional GANs which can change the shape of face components by manual editing precise masks. As shown in Figure ~\ref{ASI}(b), SC-FEGAN~\cite{Jo_2019_ICCV} requires directly taking mask, precise sketch and color as input for editing the shape of face components. However, such precise input is difficult and inconvenient to obtain.
Reference-guided methods can directly learn shape information from reference images without precise auxiliary representation, relieving the dependence on face attribute annotation or precise sketch/color/mask information for face portrait editing. Inspired by this, we propose a new framework (see Figure 2(c)), which can achieve diverse and controllable face semantic components editing (e.g., eyes, nose, mouth), which is shape free and precise landmark or sketch free.

\paragraph{Face Completion/Inpainting. } 
Face completion, also known as face inpainting, aims to complete a face image with a masked region or missing content.
Early face completion works~\cite{bertalmio2000image,criminisi2003object,bertalmio2003simultaneous} fill semantic contents based on the overall image and structural continuity between the masked and unmasked regions, which aims to reconstruct missing regions according to the ground-truth image.
Recently, some learning-based method~\cite{zheng2019pluralistic,song2018geometry} are proposed for generating multiple and diverse plausible results. ~\cite{zheng2019pluralistic} proposes a probabilistically principled framework with two parallel paths, the VAE-based reconstructive path is used to impose smooth priors for the latent space of complement regions, the generative path is used to predict the latent prior distribution for missing regions, which can be sampled to generate diverse results. However, this method lacks controllability for diverse results.
In light of this, ~\cite{song2018geometry} generate controllable diverse results from the same masked input by manual modifying facial landmark heatmaps and parsing maps.

\section{Method}
In this section, we first introduce the framework of reference-based face component editing. Then, the example-guided attention module are presented. Finally, objective functions of the proposed model are provided.

\subsection{Reference Guided Face Component Editing}
We propose a framework (See Figure~\ref{framework}), named reference guided face component editing, that transfers one or multiple face components of a reference image to corresponding components of the source image. The framework requires three inputs, a source image $\mathbf{I}_{s}$, a reference image $\mathbf{I}_{r}$ and the target face component mask of the reference image $\mathbf{I}_{m}^{s}$. The source mask $\mathbf{I}_{m}^{s}$ merely needs to roughly represent the target face components, which can be obtained by a face parsing or landmark detection network. The corrupted image can be obtained by Equation 1:
\begin{equation}
\label{E1}
\mathbf{I}_{c}=\mathbf{I}_{s}\ast\mathbf{I}_{m}^{s},
\end{equation}
where $\ast$ is an element-wise multiplication operator. The goal of this framework is to generate a photo-realistic image $\mathbf{I}_{g}$, in which shape is semantically similar to corresponding face components of the reference image while the face color and topological structure are consistent with the source image.

In this work, we utilize an image inpainting generator $\mathcal{G}_{i}$ as backbone that can generate the completed image without constraints on shape, while a discriminator $\mathcal{D}$ is used for distinguishing face images from the real and synthesized domains. $\mathcal{G}_{i}$ is consist of an encoder, seven dilated residual blocks, an attention module and a decoder. To fill missing parts with semantically meaningful face components of a reference image, an reference-guided encoder $\mathcal{E}_{r}$ is introduced to extract features of the reference image. The encoder of $\mathcal{G}_{i}$ and $\mathcal{E}_{r}$ have same structure but parameters are not shared. Attention module, to be described next, is effectively transferring semantic components from high-level features of the reference image to $\mathcal{G}_{i}$, further improving the performance of our framework. The generated image $\mathbf{I}_{g}$ can be expressed as:
\begin{equation}
\label{E2}
\mathbf{I}_{g}=\mathcal{G}_{i}(\mathbf{I}_{c},\mathbf{I}_{m}^{s},\mathcal{E}_{r}(\mathbf{I}_{r})),
\end{equation}

\subsection{Example-guided Attention Module}
Inspired by the short+long term attention of PICNet~\cite{zheng2019pluralistic}, we propose an example-guided attention. The short+long term attention uses the self-attention map to harness distant spatial context and the contextual flow to capture feature-feature context for finer-grained image inpainting. The example-guided attention replaces the contextual flow by the example-guided flow which combines the attention features and the reference features for clearly transforming the corresponding face component features of reference images to source images.

The proposed structure is shown in the upper right corner of Figure ~\ref{framework}. Following~\cite{zheng2019pluralistic}, the self-attention map is calculated from the source feature, which can be expressed as $\mathbf{F}_{a}=\mathbf{F}_{s}\otimes{\mathbf{F}_{m}}$. The attention map $\mathbf{F}_{m}$ is obtained by $\mathbf{F}_{m}=Softmax(\mathbf{F}_{q}^{T}\mathbf{F}_{q})$, in which $\mathbf{F}_{q}=Conv(\mathbf{F}_{s})$ and $Conv$ is a $1\times1$ convolution filter. To transfer the target face component features of reference images, the reference feature is embedded by multiplying the attention map with the source mask $\mathbf{I}_{m}^{s}$. The example-guided flow is expressed as follows:
\begin{equation}
\label{E12}
\mathbf{F}_{e}=\mathbf{I}_{m}^{s}*\mathbf{F}_{e}'+(1-\mathbf{I}_{m}^{s})*\mathbf{F}_{r},
\end{equation}
where $\mathbf{F}_{e}'=\mathbf{F}_{r}\otimes{\mathbf{F}_{m}}$. Finally, the fused feature $\mathbf{F}_{f}=\mathbf{F}_{a}\oplus{\mathbf{F}_{e}}$ is sent to the decoder for generating results with the target components of the reference image.

\subsection{Objective}
We use the combination of a per-pixel loss, a style loss, a perceptual loss, a contextual loss, a total variation loss and an adversarial loss for training the framework.

To capture fine facial details we adopt the perceptual loss~\cite{johnson2016perceptual} and the style loss~\cite{johnson2016perceptual}, which are widely adopted in style transfer, super resolution, and face synthesis. The perceptual loss aims to measure the similarity of the high dimensional features (e.g., overall spatial structure) between two images, while the style loss measure the similarity of styles (e.g., colors). The perceptual loss can be expressed as follows:
\begin{equation}
\label{E3}
\mathcal{L}_{perc}=\sum_{l}\frac{1}{{C_l}{H_l}{W_l}}\|\phi_{l}(\mathbf{I}_{g})-\phi_{l}(\mathbf{I}_{s})\|_{1},
\end{equation}
The style loss compare the content of two images by using Gram matrix, which can be expressed as follows:
\begin{equation}
\label{E4}
\mathcal{L}_{style}=\sum_{l}\frac{1}{{C_l}{C_l}}\|\frac{G_{l}(\mathbf{I}_{g}\ast{\mathbf{I}_{m}^{s})}-G_{l}(\mathbf{I}_{c})}{{C_l}{H_l}{W_l}}\|_{1},
\end{equation}
where $\|\cdot\|_{1}$ is the $\ell_{1}$ norm. $\phi_{l}(\cdot) \in \mathbb{R}^{{C_l}\times{H_l}\times{W_l}}$ represents the feature map of the $l$-th layer of the VGG-19 network~\cite{simonyan2014very} pretrained on the ImageNet. $G_{l}(\cdot)=\phi_{l}(\cdot)^{T}\phi_{l}(\cdot)$ represents the Gram matrix corresponding to $\phi_{l}(\cdot)$.

The contextual loss~\cite{mechrez2018contextual} measures the similarity between non-aligned images, which in our model effectively guarantees the consistent shape of the target face components in generated images and reference images. Given an image x and its target image y, each of which is represented as a collection of points(e.g. VGG-19~\cite{simonyan2014very} features): $X=\{x_{i}\}$ and $Y=\{y_{j}\}$, $|X|=|Y|=N$. The similarity between the images can be calculated that for each feature $y_{j}$, finding the feature $x_{i}$ that is most similar to it, and then sum the corresponding feature similarity values over all $y_{j}$. Formally, it is defined as:
\begin{equation}
\label{E5}
CX(x,y)=CX(X,Y)=\frac{1}{N}\sum_{j}\max_{i}CX_{ij},
\end{equation}
where $CX_{ij}$ is the similarity between features $x_{i}$ and $y_{j}$. At training stage, we need the target components mask of reference images $I_{m}^{r}$ for calculating the contextual loss. The contextual loss can be expressed as follows:
\begin{equation}
\label{E6}
\mathcal{L}_{cx}=-\log(CX(\phi_{l}(\mathbf{I}_{g}\ast{\mathbf{I}_{m}^{s}}),\phi_{l}(\mathbf{I}_{r}\ast{\mathbf{I}_{m}^{r}}))),
\end{equation}

The per-pixel loss can be expressed as follows:
\begin{equation}
\label{E7}
\mathcal{L}_{pixel}=\|\phi_{l}(\mathbf{I}_{g}\ast{\mathbf{I}_{m}^{s}})-\phi_{l}(\mathbf{I}_{s}\ast{\mathbf{I}_{m}^{s}})\|_{1},
\end{equation}

Lastly, we use an adversarial loss for minimizing the distance between the distribution of the real image and the generated image. Here, LSGAN~\cite{mao2017least} is adopted for stable training. The adversarial loss can be expressed as follows:
\begin{equation}
\label{E8}
\mathcal{L}_{ad_{G}}=\mathbb{E}[(\mathcal{D}(\mathbf{I}_{g})-1)^{2}],
\end{equation}
\begin{equation}
\label{E9}
\mathcal{L}_{ad_{D}}=\mathbb{E}[\mathcal{D}(\mathbf{I}_{g})^{2}]+\mathbb{E}[(\mathcal{D}(\mathbf{I}_{s})-1)^{2}],
\end{equation}

With total variational regularization loss $\mathcal{L}_{tv}$~\cite{johnson2016perceptual} added to encourage the spatial smoothness in the generated images, we obtain our full objective as:
\begin{equation}
\label{E10}
\begin{split}
\mathcal{L}=\lambda_{1}\mathcal{L}_{perc}+\lambda_{2}\mathcal{L}_{style}+\lambda_{3}\mathcal{L}_{cx}\\
+\lambda_{4}\mathcal{L}_{pixel}+\lambda_{5}\mathcal{L}_{tv}+\lambda_{6}\mathcal{L}_{ad_{G}}
\end{split}
\end{equation}

\section{Experiments}
\subsection{Dataset and Preprocessing}
The face attribute dataset CelebAMask-HQ~\cite{lee2019maskgan} contains 30000 aligned facial images with the size of $1024\times1024$ and corresponding 30000 semantic segmentation labels with the size of $512\times512$. Each label in the dataset has 19 classes (e.g., "left eye", "nose", "mouth"). In our experiments, three face components are considered, i.e., eyes ("left eye $\&$ right eye"), nose ("nose"), and mouth ("mouth $\&$ u$\_$lip $\&$ l$\_$lip"). We obtain rough version of face components from semantic segmentation labels by an image dilation operation, which are defined as mask images. We take 2,000 images as the test set for performance evaluation, using rest images to train our model. All images are resized to $256\times256$.
\subsection{Implementation Details}
Our end-to-end network is trained on four GeForce GTX TITAN X GPUs of 12GB memory. Adam optimizer is used in experiments with $\beta_1 = 0.0$ and $\beta_2 = 0.9$. The hyperparameters from $\lambda_{1}$ to $\lambda_{6}$ are assigned as 0.1, 250, 1, 0.5, 0.1, 0.01 respectively. For each source image, we remove two or three target face components for training. During testing, our model can change one or more face components.

\begin{figure}[!tbp]
\centering
\includegraphics[width=0.47\textwidth]
{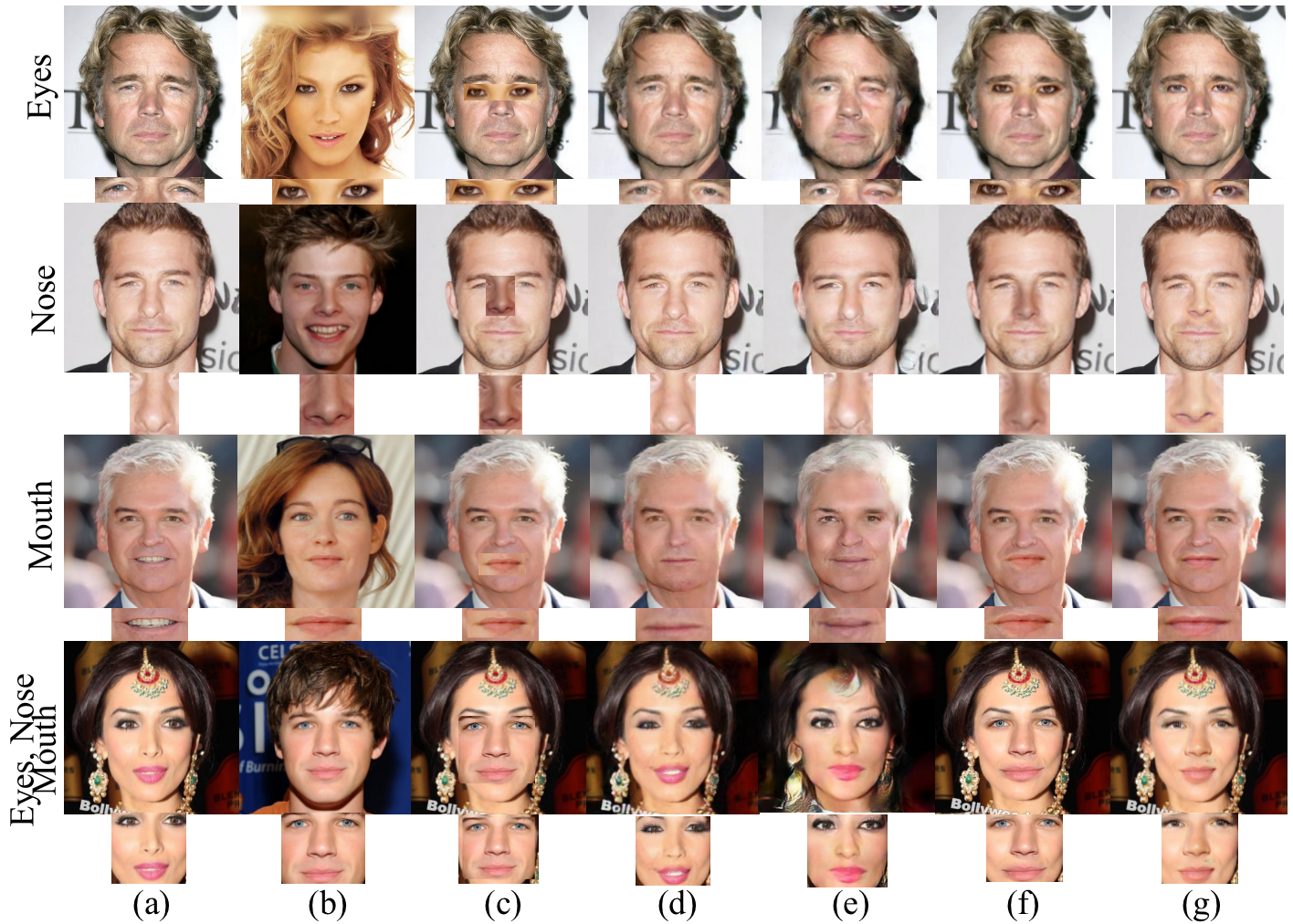}
\caption{Comparisons among (c) copy-and-paste, (d) AttGAN, (e) ELEGANT, (f) the commercial state of the art algorithm in Adobe Photoshop and (g) the proposed r-FACE technique. The source and reference images are shown in (a) and (b), respectively. AttGAN and ELEGANT edit attributes: $'Narrow\_Eyes'$, $'Pointy\_Nose'$ ,$'Mouth\_Slightly\_Open'$ and all of them respectively.}
\label{comparison}
\end{figure}
\subsection{Qualitative Evaluation}
\paragraph{Comparison with Other Methods. }  In order to demonstrate the superiority of the proposed method, we compare the quality of sample synthesis results to several benchmark methods.
In addition to face editing model AttGAN~\cite{he2019attgan} and ELEGANT~\cite{xiao2018elegant}, we also consider copy-and-paste as a naive baseline and Adobe Photoshop image editing as an interactive way to produce synthesized face images.
According to the results presented in Figure~\ref{comparison}, although margins of edited facial components in Adobe Photoshop are much smoother than those in results of copy-and-paste, obvious ghosting artifacts and color distortions still exist and more fine-grained manual labour is required to improve the quality.
\begin{figure}[!tbp]
\setlength{\belowcaptionskip}{-0.4cm}
\setlength{\abovecaptionskip}{0.15cm}
\centering
\includegraphics[width=0.44\textwidth]
{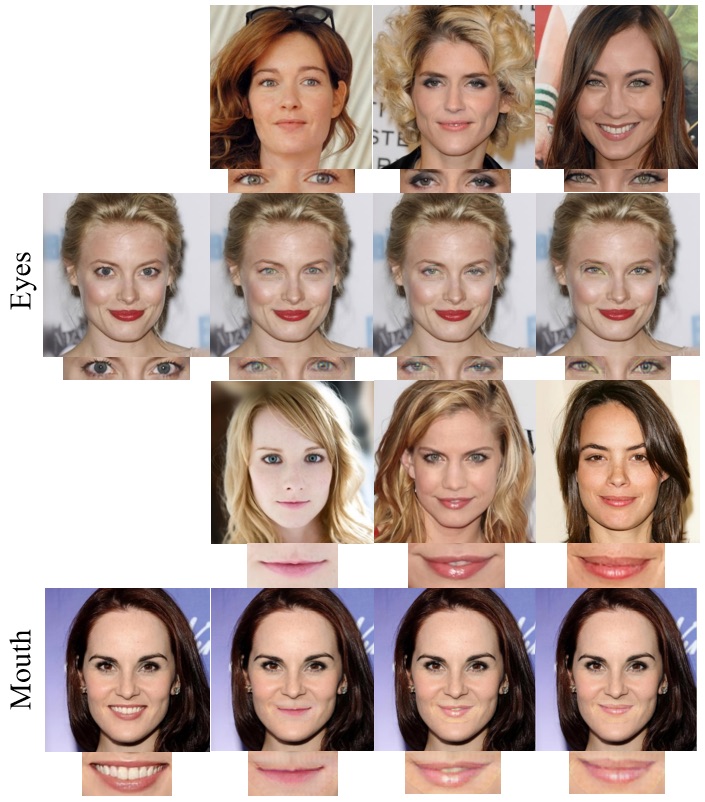}
\caption{Illustrations of multi-modal face components editing, including 'eyes' and 'mouth'. The first column represents source images, the first and third rows are reference images, and the second and fourth rows are synthesized images according to reference images above.}
\label{multi}
\end{figure}
In contrast, AttGAN could generate realistic synthetic images which are indistinguishable to generic ones in an end-to-end manner.
However, pre-defined facial attribute labels are used to guide AttGAN to transform facial components thus the diversity of generated images is limited.
Moreover, as can be seen in Figure~\ref{comparison}, AttGAN only produces subtle changes that could hardly reflect the desired translation.
ELEGANT, as a reference-guided face attribute editing method, can learn obvious semantic attributes (e.g., open eyes or close mouth), but could not learn abstract shape information (e.g., nose editing). Moreover, ELEGANT produces large deformation and unexpected artifacts at other attribute-unrelated areas, especially the editing of multiple components. 
On the contrary, our method takes arbitrary face images as reference, which significantly increases the diversity of generated images.

\paragraph{Multi-Modal Face Components Editing. }  Reference-guided face components editing improves the diversity and controllability of generated faces, as the stylistic information is designated by arbitrary reference images.
As shown in Figure~\ref{multi}, target face components of interest (e.g. eyes and mouths) are transformed to be of the same style as the corresponding reference image.
For example, synthesized mouths in faces of the last row accurately simulate the counterpart in reference images, in terms of both overall shape (e.g. raised corners of pursed mouth) and local details (e.g. partially covered teeth).
Meanwhile, they are naturally blended in to the source face without observable color distortions and ghosting artifacts, demonstrating the effectiveness of proposed method.

\subsection{Quantitative Evaluation}
Following most of face portrait methods~\cite{he2019attgan,wu2019relgan}, we leverage Fr\'{e}chet Inception Distance (FID, lower value indicates better quality)~\cite{heusel2017gans} and Multi-Scale Structural SIMilarity (MS-SSIM, higher value indicates better quality)~\cite{wang2003multiscale} to evaluate the performance of our model. FID is used to measure the quality and diversity of generated images. MS-SSIM is used to evaluate the similarity of two images from luminance, contrast, and structure.

We compare our method with AttGAN\cite{he2019attgan}, one of label-conditioned methods, and ELEGANT~\cite{xiao2018elegant}, one of reference-guided methods. For AttGAN and ELEGANT, the value of FID or MS-SSIM are the result of averaging three pre-defined attributes for shape changing, including 'Narrow\_Eyes', 'Pointy\_Nose' and 'Mouth\_Slightly\_Open'.
The backbone of r-FACE is similar to image inpainting task, so we also compare our method with GLCIC\cite{iizuka2017globally}, one of popular face inpainting methods.
As is shown in TABLE~\ref{m1}, comparing with these methods, the FID of our method is much lower.
With the observation that the MS-SSIM of our method is lower than AttGAN and GLCIC, we analyze the reasons: MS-SSIM is sensitive to luminance, contrast, and structure, however, 1) GLCIC does not have any constraints on the structure or shape of components, just requiring the missing regions to be completed; 2) AttGAN edits shape-changing attributes with subtle changes that are hard to be observed as shown in Figure ~\ref{comparison}, so the change of luminance, contrast, and structure is limited. In contrast, r-FACE imposes a geometric similarity constraint on the components of source images and reference images, which changes the shape or structure drastically and even affects the identity of the face.

\begin{table}[!tbp]
\centering
	\begin{tabular}{ccccccc}
		\toprule
		Method &  FID $\downarrow$ & MS-SSIM $\uparrow$\\
		\midrule
		GLCIC\cite{iizuka2017globally} & 8.09 & 0.95 \\
		AttGAN\cite{he2019attgan} & 6.28 & 0.96 \\
		ELEGANT~\cite{xiao2018elegant} & 15.97 & 0.82\\
        \midrule
        \textbf{r-FACE (Ours)} & \textbf{5.81} & \textbf{0.92} \\
		w / o attention & 6.25 & 0.90 \\
        w / o contextual loss & 5.27 & 0.90 \\
        w / o style loss & 8.28 & 0.90 \\
        w / o perceptual loss & 8.49 & 0.89 \\
		\bottomrule
	\end{tabular}
\setlength{\belowcaptionskip}{-0.4cm}
\caption{Comparisons of FID and MS-SSIM on the CelebA-HQ dataset.}
\label{m1}
\end{table}
\subsection{Ablation Study}
\paragraph{Example-guided Attention Module.} 
To investigate the effectiveness of the example-guided attention module, we conduct a variant of our model, denoted as ”r-FACE w / o attention”. In ”r-FACE w / o attention”, we train r-FACE without example-guided attention module. To learn face components from a reference image, we directly concatenate the features of reference images with that of source images, which introduced the features of the whole reference image. TABLE~\ref{m1} shows the comparison on FID and MS-SSIM between "r-FACE w / o attention" and "r-FACE". Our method outperforms "r-FACE w / o attention" on both FID and MS-SSIM, which indicates that example-guided attention module can explicitly transfer the corresponding face components of a reference image and reduce the impact of other information of the reference image.

\begin{figure}[!tbp]
\setlength{\belowcaptionskip}{-0.4cm}
\setlength{\abovecaptionskip}{0.15cm}
\centering
\includegraphics[width=0.45\textwidth]
{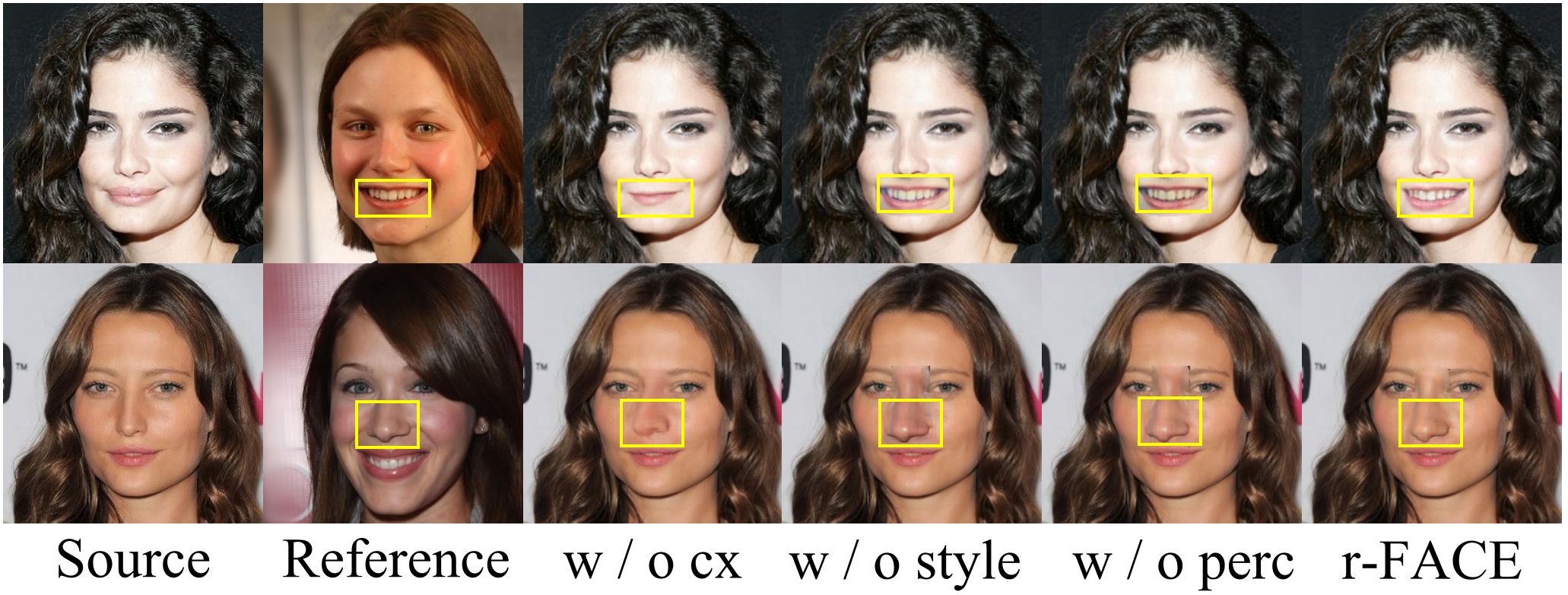}
\caption{Visual comparisons for different loss configurations, including "r-FACE w / o contextual loss", "r-FACE w / o style loss", "r-FACE w / o perceptual loss" and "r-FACE".}
\label{ablation}
\end{figure}

\paragraph{Loss Configurations.}
In this section, we conduct ablation studies on different loss, such as the contextual loss, the style loss and the perceptual loss, for evaluating their individual contributions on our framework. As shown in TABLE~\ref{m1}, the quantitative comparison among "r-FACE w / o contextual loss", "r-FACE w / o style loss", "r-FACE w / o perceptual loss" and "r-FACE" are presented. "r-FACE" outperforms other loss configurations on MS-SSIM, which indicates the effectiveness of the three losses. It is observed that the FID of "r-FACE" is also better than other loss configurations except "r-FACE w/o contextual loss". We argue that contextual loss plays an important role in restricting the shape of face components. The framework has no constraints on the shape after removing the contextual loss, so "r-FACE w / o contextual loss" is easier to generate missing components, merely requiring generated images to look real. Figure ~\ref{ablation} compares the visual effects of different loss configurations. We find "r-FACE w / o contextual loss" could not synthesize components with the corresponding shape of reference images, which indicates the contextual loss is significantly important in shape constraints.
In the results of "r-FACE w / o style loss" and "r-FACE w / o perceptual loss", color distortions and obvious ghosting artifacts are observed, which show the style loss and the perceptual loss are able to preserve skin color of source image.
Above all, we prove that three losses contribute to the performance of our framework and the combination of all losses achieves the best results.

\subsection{Discussion and Limitation}
Reference guided face component editing has wide real-life applications in interactive entertainment, security and data augmentation. Given whole reference images of any identity, r-FACE performs various face component editing, which can achieve the effect of "plastic surgery". Besides, when all face components of a reference face are transformed to the source face, r-FACE is further extended to face swapping task. Moreover, As the face component is a part of face, editing face components may change the identity of the person. Therefore, r-FACE can be used for data augmentation to generate face images of different identities. Although r-FACE obtains diverse and controllable face component editing results, reference images with significant differences in pose are still challenging for our model. We will continue to explore solving extreme pose problems in further work.

\section{Conclusion}
In this work, we have proposed a novel framework, reference guided face components editing (r-FACE), for high-level face components editing, such as eyes, nose and mouth. r-FACE can achieve diverse, high-quality, and controllable component changes in shape from given references, which breaks the shape and precise intermediate presentation limitation of existing methods. For embedding the target face components of reference images to source images specifically, an example-guided attention module is designed to fuse the features of source images and reference images, further boosting the performance of face component editing. The extensive experiments demonstrate that our framework can achieve state-of-art face editing results with observable geometric changes.

\section*{Acknowledgments}

This work was partially supported by the Natural Science Foundation of China (Grant No. U1836217, Grant No. 61702513, Grant No. 61721004, and Grant No. 61427811). This research was also supported by Meituan-Dianping Group, CAS-AIR and Shandong Provincial Key Research and Development Program (Major Scientific and Technological Innovation Project) (NO.2019JZZY010119).

\bibliographystyle{named}
\end{document}